\documentclass{bmvc2k}

\title{Depth-only Object Tracking}

\addauthor{Song Yan}{song.yan@tuni.fi}{1}
\addauthor{Jinyu Yang}{jinyu.yang96@outlook.com}{23}
\addauthor{Ale{\v{s}} Leonardis}{a.leonardis@cs.bham.ac.uk}{3}
\addauthor{Joni-Kristian K\"am\"ar\"ainen}{joni.kamarainen@tuni.fi}{1}

\addinstitution{Tampere University\\ Finland}
\addinstitution{Southern University of Science and Technology\\China}
\addinstitution{University of Birmingham\\United Kingdom}

\runninghead{~S. Yan, J. Yang, A. Leonardis, J.-K. K\"am\"ar\"ainen}{Depth-only Object Tracking~}

\def\eg{\emph{e.g}\bmvaOneDot}

\usepackage{graphics}
\usepackage{booktabs}
\usepackage{paralist} % compactitem
\usepackage{amssymb}
\usepackage{bbold}
\usepackage{amsmath}

\begin{document}

\maketitle

\begin{abstract}
Depth (D) indicates occlusion and is less sensitive to illumination changes, which make depth attractive modality for Visual Object Tracking (VOT).
Depth is used in RGBD object tracking where the best trackers are deep RGB trackers with additional heuristic using depth maps. 
There are two potential reasons for the heuristics: 
1) the lack of large RGBD tracking datasets to train deep RGBD trackers and 
2) the long-term evaluation protocol of VOT RGBD that benefits from heuristics such as depth-based occlusion detection. 
In this work, we study how far D-only tracking can go if trained with large amounts of depth data. 
To compensate the lack of depth data, we generate depth maps for tracking. 
We train a "Depth-DiMP" from the scratch with the generated data and fine-tune it with the available small RGBD tracking datasets.
The depth-only DiMP achieves good accuracy in depth-only tracking and combined with the original RGB DiMP the end-to-end trained RGBD-DiMP outperforms the recent VOT 2020 RGBD winners.
\end{abstract}
%---------------------------------------------------------------------------
%---------------------------------------------------------------------------

\section{Introduction}
\label{sec:intro}
The Visual Object Tracking (VOT) Challenge has been running since 2013. 
In 2019 a new challenge, VOT-RGBD, and dataset for RGBD tracking were introduced. 
Interestingly, the winners of VOT-RGBD 2019~\cite{Vot2019} and 2020~\cite{Vot2020} are
actually RGB-only trackers (MDNet~\cite{MDNet}, ATOM~\cite{ATOM} and DiMP~\cite{DiMP})
with additional heuristics using the D channel.
On one hand, VOT-RGBD is a long-term tracking challenge that benefits from heuristic
full occlusion detection, and  
on the other hand, the lack of large RGBD tracking datasets makes it impossible to train RGBD trackers from scratch.
It remains unclear to what degree depth is useful in object tracking.
%and ultimately what the tracking scenes where depth is clearly superior to visual information?

In this work, we aim to answer the above research question by training a deep depth-only tracker from scratch using depth tracking data and evaluating it with the previous VOT-RGBD datasets
(CDTB~\cite{CDTB} in 2019-2021 and DepthTrack~\cite{DepthTrack} in 2021).
This work provides the following novel contributions:
{\bf 1)} we introduce synthetic depth-only tracking datasets that are generated from the large RGB datasets using monocular depth estimation;
{\bf 2)} we introduce a depth-only DiMP (Depth-DiMP) and its extension RGBD-DiMP which are trained using both generated and real data.
With the help of Depth-DiMP, 
{\bf 3)} we report a careful analysis of depth-only tracking and compare the performance of RGB and D-only trackers on CDTB and DepthTrack datasets. 
Interestingly, our Depth-DiMP is almost on par with the best RGB trackers and RGBD-DiMP wins the VOT-RGBD winners on the DepthTrack test set.
Our analysis shows that there are scenes for which depth is an important complementary cue for object tracking.
%---------------------------------------------------------------------------
%---------------------------------------------------------------------------

\section{Related work}
\paragraph{Visual object tracking.}
In recent years, there has been steady progress from conventional "engineered" trackers toward learning-based deep trackers. 
Discriminative Correlation Filter (DCF)~\cite{Henriques_KCF} is a representative example of the conventional trackers. 
DCF formulation leads to optimized detection filters that
are used to find the target in the next frame. This principled
idea has inspired the main architecture of the two well-performing
deep trackers: ATOM~\cite{ATOM} and
DiMP~\cite{DiMP}. Their architectures have three
processing modules:
i) feature extraction,
ii) target classifier and
iii) target estimation.
For the depth-only D and RGBD trackers in this work
we adopt the DiMP architecture.
A DiMP extension, Probabilistic DiMP (PrDiMP)~\cite{PrDiMP}, has been proposed, but we adopted
the original DiMP due to its simplicity and stable public implementation.

\vspace{-2\medskipamount}

\paragraph{RGBD and D Object Tracking.}
RGBD tracking has been a separate track in the VOT Challenge since 2019~\cite{Vot2019}. 
The three best trackers in the VOT2020-RGBD challenge~\cite{Vot2020}, ATCAIS,  DDiMP and CLGSD, are RGBD extensions of the deep RGB trackers,  ATOM, DiMP and MDNet~\cite{MDNet}.
%Similar to 2019 RGBD trackers~\cite{Kart-2019-cvpr, DAL}, 
They use depth channel to support the RGB trackers with heuristic processing. 
Our work differs substantially from all previous RGBD trackers in the sense that our tracker is 
data-driven similar to the original ATOM and DiMP. 
To the best of authors' knowledge 
there are no other works addressing the generic object tracking with depth only.

\vspace{-2\medskipamount}

\paragraph{RGBD tracking datasets.}
Deep RGBD tracker research suffers from the
%One reason to the lack of learning-based deep D and RGBD trackers is the
fact that there are no high quality and large RGBD tracking datasets. The only available datasets are PTB, STC and CDTB. 
PTB~\cite{princetonrgbd} was perhaps the first published RGBD tracking benchmark and contains 100 video sequences, but the groundtruth is not publicly available. 
STC~\cite{STC} is more recent and of better quality, but contains only 36 sequences. 
The largest publicly available dataset is CDTB~\cite{CDTB} that contains 80 indoor and outdoor sequences captured with multiple different sensors. 
CDTB is also used in the VOT-RGBD 2019-2021 evaluations~\cite{Vot2019, Vot2020, Vot2021}.
VOT2021-RGBD provides an additional dataset, DepthTrack~\cite{DepthTrack}, from which we use
%~\footnote{At the moment of writing this paper, DepthTrack only releases 120 sequences.},
120 sequences divided to training (70) and test (50) sets. 
% its training (70) and test (50) sets.
Cross-dataset comparisons are made between CDTB and DepthTrack.
%---------------------------------------------------------------------------
%---------------------------------------------------------------------------

\section{Methods}
\label{sec:method}
In this section, we briefly describe the original DiMP and then introduce its depth-only, Depth-DiMP, and RGBD variants that differ only in their feature extraction parts.

\subsection{DiMP}
The ATOM~\cite{ATOM} and DiMP~\cite{DiMP} trackers share the same overall architecture: 
1) a pre-trained feature extraction module, 
2) an offline trained target estimation module and 
3) an online trained classifier module.

\paragraph{Feature extraction.}
DiMP employs the ResNet~\cite{resnet} network as its feature extraction module. 
The deeper ResNet-50 provides better performance, but is slower than the lighter ResNet-18 (43 vs. 57 fps). 
We use ResNet-50 features extracted from the same layers as in the original DiMP. 
ResNet is pre-trained with the ImageNet image categorization dataset~\cite{deng2009imagenet} to learn strong semantic features. 
Ideally, ResNet should be trained with D and RGBD depth data for the D and RGBD trackers, 
but since there are no large semantic tasks depth datasets, 
we utilize RGB-trained ResNet features as the initial features for both RGB and depth.

\paragraph{Target classifier.}
The first step in the DiMP processing pipeline is to estimate
the target location in the current frame given its location in
the previous frame ({\em reference frame}). 
The target location is estimated by an online trained {\em target classifier} module. 
The ATOM classifier module is a simple two-layer fully connected regressor network,
\begin{equation}
 f(x; w) = \phi_2(w_2*\phi_1(w_1*x))   \enspace 
\end{equation}
where $w=\left\{w_1,w_2\right\}$ are the network weights,
$*$ is the multi-channel convolution, 
$\phi_s$ are activation functions and $x$ is the ResNet
feature map. During tracking negative and positive examples are
generated from the reference frame and the weights are re-optimized over
a small number of iterations using the Gauss-Newton method~\cite{nocedal2006numerical}.
It is noteworthy that this step is similar to
the DCF formulation where $w_2$ resembles the discriminative
correlation filters for tracking. DiMP takes this step further
by introducing a more complex and powerful classifier network.
DiMP optimization is based on the regularized loss
\begin{equation} \label{eq:dimp_loss}
        L(f) = \frac{1}{|S_{train|}} \sum_{(x, c)\in S_{train}} \parallel	r (x \circledast f, c) \parallel^2 + \parallel \lambda f \parallel^2 
    \enspace 
\end{equation}
where $f$ are the filter weights to be optimized and $c$ denotes the desired output response for the randomly selected training samples $S_{train}$. 
$\circledast$ denotes convolution and $\lambda$ is a regularization factor.
Suitable candidates for $c$ are a Gaussian peaked at the groundtruth location or a triangular basis function of DiMP. 
DiMP defines an error residual $r(\cdot,\cdot)$ that combines least-squares and the hinge loss 
and is optimized using the Gauss-Newton method.

\paragraph{Target estimation.}
The target estimation networks of ATOM and DiMP share the same structure.
Given the position estimated from the classifier module at the maximum score of the response $x \circledast f$ 
and the bounding box width and height from the previous frame 
10 bounding box proposals are generated by adding uniform random noise. 
The target estimation module is applied to each proposal. 
The final estimate is the average of three best proposals after the target estimation stage. 
The target estimation provides Intersection-over-Union (IoU) score for each proposal $B$ using
\begin{equation}
    IoU(B)=g(c(x_0,B_0) , z(x,B)) \enspace 
\end{equation}
where $g$ is the predictor network. The predictor network is trained offline with the largest available
RGB tracking datasets (\eg LaSOT~\cite{fan2019lasot} and TrackingNet\cite{muller2018trackingnet}) and with augmented examples from the diverse set of classes in COCO~\cite{lin2014microsoft}. 
$c(\cdot,\cdot)$ is the feature representation of the initial region $B_0$ from the feature map $x_0$ of reference image
while $z(\cdot,\cdot)$ is the proposal representation from the feature map $x$ of the test image. 

\subsection{Depth DiMP}
In all experiments we wanted to keep the number of parameters (weights) of the original DiMP fixed so that we could compare the effect of training data only. For different inputs, RGB, D and RGBD, this was achieved by altering only the ResNet-50 based feature extraction module of DiMP. These changes for D and RGBD are briefly explained next. 

\paragraph{Depth-DiMP.} 
The input affects only the feature extraction module of DiMP, i.e. ResNet-50 features extracted from the input. 
The depth values can be encoded to a single channel, but to keep the DiMP structures unchanged and the same total number of parameters (weights) we kept all three channels and experimented with different encoding of the depth. 
The two encoding experimented where {\em channel copying} and {\em colormap encoding}. 
The channel copying is simply repeating the same normalized depth map three times and is denoted by $3\times D$. 
The colormap encoding is performed by mapping pixels of the normalized depth map to RGB vectors of a predefined color matrix (JET used in the experiments).
This encoding is denoted by "ColMap".
Depth normalization is performed by normalizing all values to $[0,1]$ and before normalization metric depth values are clipped to the range $[0.0, 10.0]$ meters.

\paragraph{RGBD DiMP.}
\begin{figure}[t]
    \centering
    \includegraphics[width=0.73\linewidth]{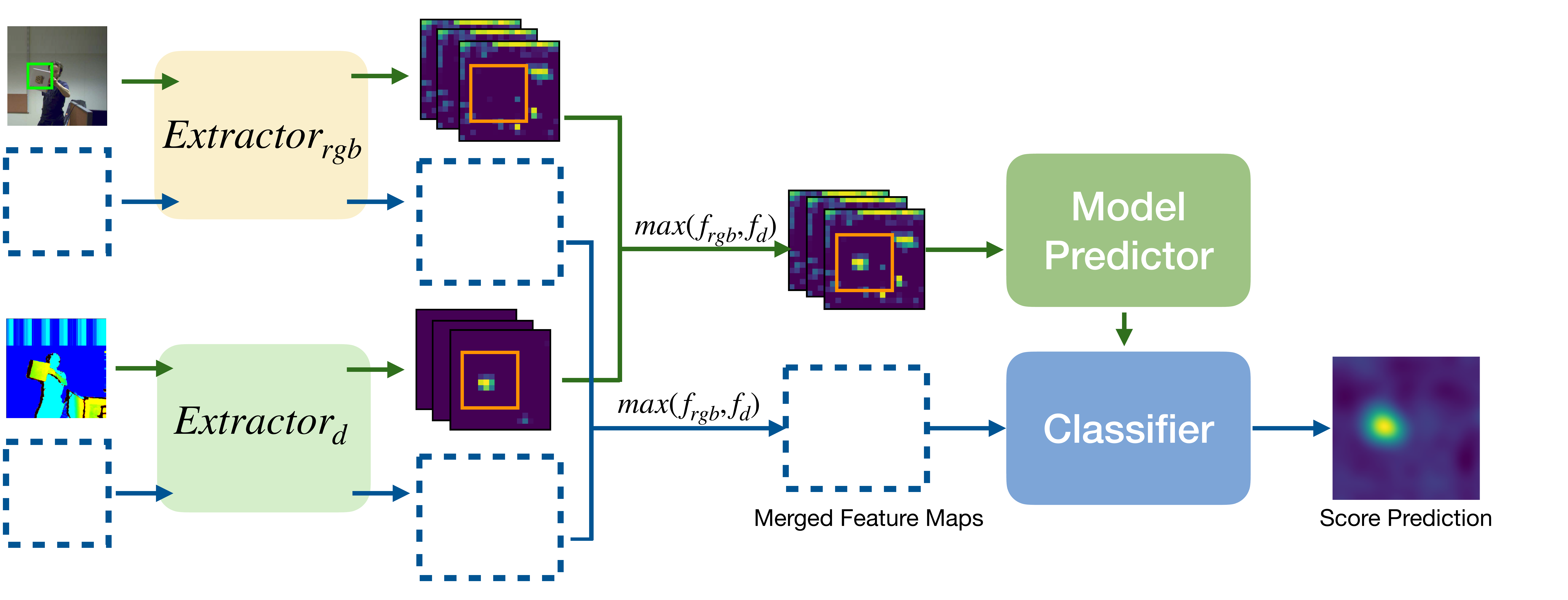}
    \caption{
    % RGBD-DiMP outline. 
    The RGBD-DiMP framework.
    $max(f_{rgb},f_d)$ denotes element-wise maximum pooling that combines the two modalities, RGB and D, before the DiMP predictor and classifier modules. To illustrate pooling, high response depth features merged with RGB features are pinpointed with orange boxes. The dashed blue boxes denote the reference branch.
    }
    \label{fig:two_path_network}
\end{figure}

To verify that RGBD tracking benefits from learning more realistic deep depth features,
we combined RGB DiMP with Depth-DiMP by adding an additional feature extraction module for the depth input.
Before the other DiMP modules the RGB and D streams are merged.
For merging, we adopt the element-wise maximum pooling. 
The RGBD-DiMP framework and merging of the two channels are illustrated in Figure~\ref{fig:two_path_network}.
%---------------------------------------------------------------------------
%---------------------------------------------------------------------------

\section{Experiments}
\label{sec:experiments}

\subsection{Performance metrics and settings}

The major difference between long- and short-term trackers is the re-detection mechanism 
and the most common way is just to brute-force search the previous location with an increasing region until the target is found.
To focus on how generic object tracking benefits from depth,
we converted the long-term (LT) sequences into short-term (ST) ones by using two rules to cut the sequences:
{\bf i)} all frames from the beginning (frame \#1) are kept until the target disappears (frame \#D) and re-appears again (frame \#D+R-1);
and {\bf ii)} a new ST sequence is started (frame \#D+R+S) when the target size is at least 75\% of its size in the frame \#1. 
This procedure is iterated until the end of each LT clip.  
However, to report numbers consistently with the previous RGBD works we adopted the LT
evaluation protocol of the  VOT challenge.
For each frame, the tracker predicts a target {\em confidence score} and {\em bounding box}.
The bounding box coordinates are used to calculate the tracking precision ({\em Pr}) by measuring the overlap ratio with the groundtruth box.
Tracking recall ({\em Re}) measures the fraction of the frames labelled as "target present" and for which the confidence score is above a threshold.
The overall evaluation {\em F-score} is the harmonic mean of {\em Pr} and {\em Re}. 
% For training the default parameter values of DiMP were used.
For training the default settings in \cite{DiMP} were used.

\subsection{Datasets}
\paragraph{RGBD datasets for evaluation.}
For the depth-only (D) and RGBD experiments we used the two largest and most recent RGBD datasets:
CDTB~\cite{CDTB} and DepthTrack~\cite{DepthTrack}.
CDTB was used in the VOT-RGBD challenge from 2019 to 2021 and DepthTrack was introduced as an additional dataset to the 2021 challenge.
Our LT-to-ST procedure produced 152 ST sequences from the original 80 CDTB LT sequences. 
Similarly, the DepthTrack 50 LT test sequences resulted to 124 ST sequences.

The 70 LT training sequences of DepthTrack were used in training as is as the training part does not differ for LT and ST.
For the cross-dataset evaluations we trained with the DepthTrack training set and tested with CDTB-ST.

\paragraph{Generated depth data for training.}
The original DiMP was trained with the training sets of
TrackingNet~\cite{muller2018trackingnet}, LaSOT~\cite{fan2019lasot}, GOT10k~\cite{huang2019got} and COCO~\cite{lin2014microsoft}. TrackingNet is the largest of these datasets including 30,000 sequences. 
The largest existing RGBD datasets are PTB, STC, CDTB and DepthTrack
of which DepthTrack is the largest containing 120 sequences.
Collecting and annotating more RGBD sequences is time consuming and tedious, and 
therefore we decided to convert the RGB datasets to RGBD using
monocular depth estimation. Depth data has much less variation than RGB and therefore we converted the two smaller datasets, LaSOT~\cite{fan2019lasot} (1,500 seqs.) and Got10k~\cite{huang2019got} (10,000 seqs.).
We also used the COCO images~\cite{lin2014microsoft} to train the tracker to have a richer set of objects. 

\begin{figure}[t]
    \centering
    \includegraphics[width=0.75\textwidth]{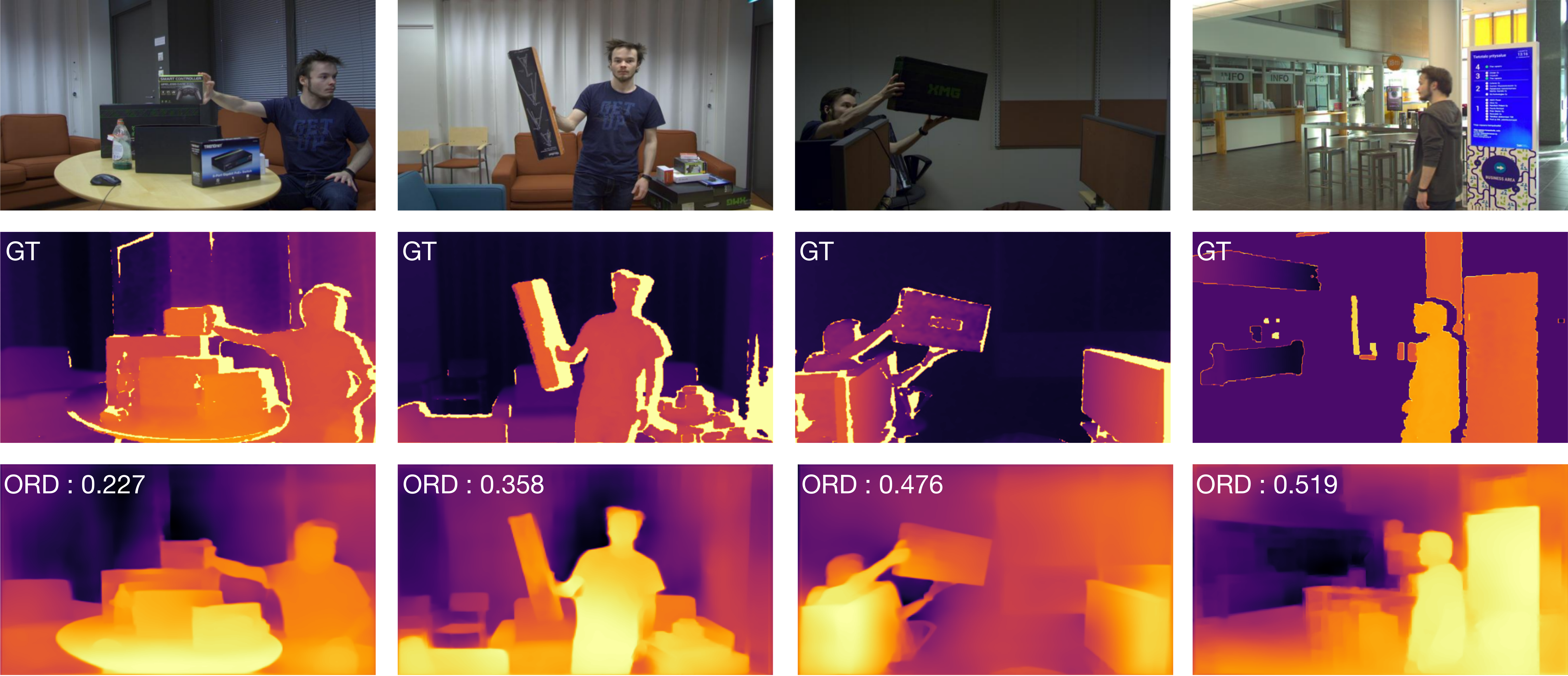}
    \caption{CDTB RGB frames (top), groundtruth depth maps (middle) and DenseDepth estimated depth maps (bottom) with their ordinal error (ORD) values. }
    \label{fig:generated_depth}
\end{figure}

\begin{figure}
    \centering
    \includegraphics[width=0.75\textwidth]{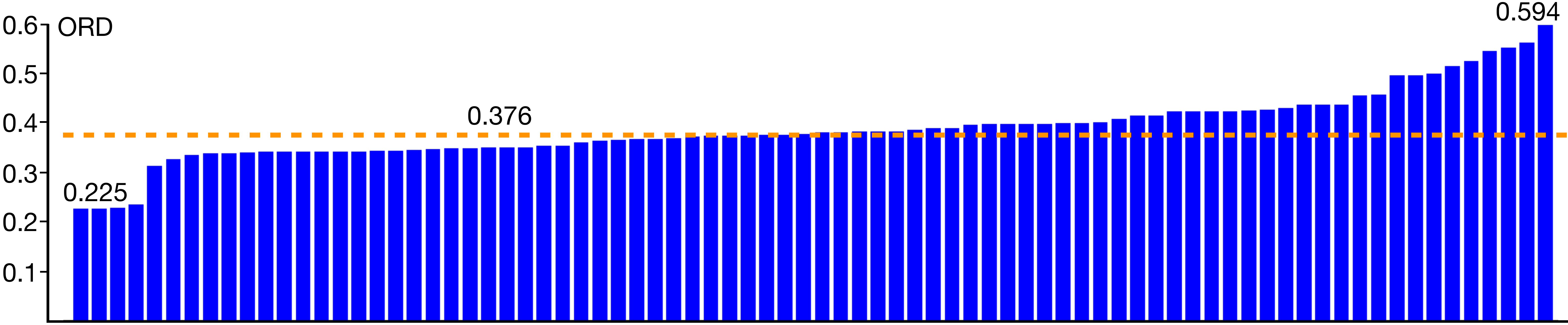}
    \caption{ORD histogram of 80 CDTB sequences. Lower ORD is better. Minimum, maximum and median values are marked. Mean ORD is 0.386.}
    \label{fig:ordinal_error_cdtb}
\end{figure}

To quantitatively evaluate the generated depth sequences
we adopted the {\em Ordinal Error} (ORD) from monocular depth literature~\cite{chen2016single, xian2020structure}. Monocular depth methods do not produce metric depth but values in $[0,1]$, and therefore evaluation is based on whether depth order of pixels match with groundtruth order. Ordinal error is defined as
\begin{equation}
  ORD = \frac{\sum_i \omega_i \mathbb{1} (\ell_{i,\tau} \neq \ell^*_{i,\tau})}{\sum_i \omega_i},
  \:\:\: where \:\:\: 
  \ell = \begin{cases}
            +1, & \text{$p_0/p_1 \geq 1+\tau$} \\
            -1, & \text{$p_0/p_1 \leq \frac{1}{1+\tau}$} \\
            ~~~0, & \text{otherwise}
          \end{cases}
    \enspace 
\end{equation} 
Computation is based on 50,000 randomly selected depth pixel pairs
$[p_0, p_1]$ from groundtruth and monocular estimated depth maps. If their ratio is greater or smaller than the thresholds defined by the parameter $\tau$ ($\tau=0.03$ adopted from literature), then positive or negative indicator values
$\ell$ (groundtruth) and $\ell^*$ (monocular estimated) are produced. ORD computes the proportion of pixel pairs that do not match. ORD $0.0$ is the ideal value and $1.0$ means that all pixels are in opposite order (for $\omega_i=1$).
We tested a number of monocular depth estimation methods and found the DenseDepth~\cite{alhashim2018high} to produce the best results.
Generated depth maps are shown in Figure~\ref{fig:generated_depth} for various ORD values and the ORD value histogram of all CDTB sequences are shown in Figure~\ref{fig:ordinal_error_cdtb}. It should be noted that depth values are missing from large parts of the groundtruth images, but otherwise the median/average ORD values 0.376/0.386 indicate sufficient quality.
Moreover, the generated RGBD sequences were manually verified and about 800 sequences were removed.

%-----------------Generated Depth ----------------------------------
% ------------------------------------------------------------------

%----------------ORD error for generated depth----------------------
%-----------------------------------------------------------------

\subsection{Cross-dataset evaluation} 
\label{sec:exp1}

\begin{table}[t]
    \centering
    \resizebox{0.8\linewidth}{!}{
    \begin{tabular}{r r r | c c c c c}
        \toprule
Pr & Re & F      & Backbone     & Pre-trained & Fine-tuned & Input & Coding\\
        \midrule
0.659 & 0.671 & 0.665 & ResNet50-RGB & RGB  &                & RGB & RGB\\ 
0.681 & 0.689 & 0.685 & ResNet50-RGB & RGB  & DepthTrack-RGB & RGB & RGB \\
\midrule
        \multicolumn{4}{c}{{\em Off-the-shelf RGB DiMP}}\\
0.625 & 0.624 & 0.625  & ResNet50-RGB & RGB  & & R  & 3$\times$R\\
0.640 & 0.637 & 0.638  & ResNet50-RGB & RGB  & & G  & 3$\times$G\\
0.624 & 0.622 & 0.623  & ResNet50-RGB & RGB  & & B  & 3$\times$B\\
0.443 & 0.402 & 0.421  & ResNet50-RGB & RGB  & & \fbox{D}  & 3$\times$D\\
0.476 & 0.431 & 0.452  & ResNet50-RGB & RGB  & & \fbox{D}  & ColMap\\
         \midrule
        \multicolumn{4}{c}{{\em + fine-tuning with real depth data}}\\
%0.659 & 0.668 & 0.663 & ResNet50-RGB & RGB  & VOT2021-RGBD-R & R & 3$\times$R\\
0.663 & 0.680 & 0.672  & ResNet50-RGB & RGB  & DepthTrack-G & G & 3$\times$G\\
%0.644 & 0.657 & 0.651 & ResNet50-RGB & RGB  & VOT2021-RGBD-B & B & 3$\times$B\\
0.510 & 0.506 & 0.508  & ResNet50-RGB & RGB  & DepthTrack-D & \fbox{D} & 3$\times$D\\
0.502 & 0.513 & 0.507  & ResNet50-RGB & RGB  & DepthTrack-D & \fbox{D} & ColMap\\
0.444 & 0.410 & 0.426  & ResNet50-RGB & RGB  & DepthTrack-D$^\dagger$ & \fbox{D} & 3$\times$D \\
0.455 & 0.412 & 0.433  & ResNet50-RGB & RGB  & DepthTrack-D$^\dagger$ & \fbox{D} & ColMap \\
%0.419 & 0.409 & 0.414 & \bf{CDTB-LT} & DiMP50 & \checkmark &  &    &   & \checkmark \\
%         \midrule
%0.501 & 0.453 & 0.476 & ResNet50-RGB & VOT2021-RGBD-D  & & D & 3$\times$D\\
%0.524 & 0.513 & 0.519 & ResNet50-RGB & VOT2021-RGBD-D  & & D & ColMap\\
         \midrule
\multicolumn{4}{c}{{\em Depth-only DiMP trained w/ generated depth}}\\
%-     & -     & -     & ResNet50-RGB & R    & VOT2021-RGBD-R & R & 3$\times$R\\
0.701 & 0.701 & 0.701  & ResNet50-RGB & G    & DepthTrack-G & G & 3$\times$G\\
0.698 & 0.703 & 0.700  & ResNet50-RGB & G    & DepthTrack-G & G & ColMap\\
%-     & -     & -     & ResNet50-RGB & B    & VOT2021-RGBD-B & B & 3$\times$B\\
%-     & -     & -     & ResNet50-RGB & D    & VOT2021-RGBD-D & D & 3$\times$D\\
0.487 & 0.375 & 0.424  & ResNet50-RGB & G    & DepthTrack-D & \fbox{D} & 3$\times$D\\
% 0.524 & 0.513 & 0.519 & ResNet50-RGB &     & VOT2021-RGBD-D  & D & 3$\times$D\\
0.549 & 0.563 & {\bf 0.556}  & ResNet50-RGB & D$^\dagger$    & DepthTrack-D & \fbox{D} & 3$\times$D\\
0.520 & 0.529 & \underline{0.525}  & ResNet50-RGB & D$^\dagger$    & DepthTrack-D & \fbox{D} & ColMap\\
%0.438 & 0.401 & 0.419 & \bf{CDTB-LT} & ResNet50 & \checkmark &            &            &            & \checkmark \\
         \bottomrule
    \end{tabular}}
    \caption{Depth-only tracking results for the CDTB-ST (ST: short-term) dataset. The results are cross-dataset where CDTB was only used in testing. 
    D$^\dagger$ denotes DenseDepth generated depth data.
    The best depth-only (\fbox{D}) F-score is bolded and the second best underlined.}
    \label{tab:exp_depth}
\end{table}
%
%---------------------------------------------------------------
%
\begin{table}[!htb]
    \centering
    \resizebox{0.78\linewidth}{!}{
    \begin{tabular}{r r r | c c c c c}
        \toprule
Pr & Re & F           & Backbone     & Pre-trained & Fine-tuned & Input & Coding\\
        \midrule
% 0.528 & 0.477 & 0.501 & ResNet50-RGB & RGB  &            & RGB & RGB\\ 
0.534 & 0.483 & 0.507 & ResNet50-RGB & RGB  &            & RGB & RGB\\ 
0.542 & 0.488 & 0.513 & ResNet50-RGB & RGB  & CDTB-RGB   & RGB & RGB \\
\midrule
        \multicolumn{4}{c}{{\em Off-the-shelf RGB DiMP}}\\
0.461 & 0.387 & 0.420 & ResNet50-RGB & RGB  &            & G   & 3$\times$G\\
0.350 & 0.246 & 0.288 & ResNet50-RGB & RGB  &            & \fbox{D}   & 3$\times$D\\
0.362 & 0.267 & 0.307 & ResNet50-RGB & RGB  &            & \fbox{D}   & ColMap\\
\midrule
        \multicolumn{4}{c}{{\em + fine-tuning with real depth data}}\\
0.463 & 0.413 & 0.437      & ResNet50-RGB & RGB  & CDTB-G           & G        & 3$\times$G\\
0.412 & 0.335 & \bf{0.370} & ResNet50-RGB & RGB  & CDTB-D           & \fbox{D} & 3$\times$D\\
0.381 & 0.350 & 0.364      & ResNet50-RGB & RGB  & CDTB-D           & \fbox{D} & ColMap\\
0.386 & 0.306 & 0.341      & ResNet50-RGB & RGB  & CDTB-D$^\dagger$ & \fbox{D} & 3$\times$D\\
0.390 & 0.328 & 0.356      & ResNet50-RGB & RGB  & CDTB-D$^\dagger$ & \fbox{D} & ColMap\\
\midrule
\multicolumn{4}{c}{{\em Depth-only DiMP trained w/ generated depth}}\\
%0.459 & 0.263 & 0.333 & ResNet50-RGB & CDTB-D & & D & 3$\times$D\\
%0.433 & 0.254 & 0.320 & ResNet50-RGB & CDTB-D & & D & ColMap \\
%0.429 & 0.239 & 0.307 & ResNet50-RGB & CDTB-D (ST) & & D & 3$\times$D \\
%0.467 & 0.231 & 0.309 & ResNet50-RGB & CDTB-D (ST) & & D & ColMap \\
% 0.338 & 0.296 & 0.315 & ResNet50-RGB & D$^\dagger$  &        & D & 3$\times$D \\
% 0.337 & 0.268 & 0.299 & ResNet50-RGB & D$^\dagger$  &        & D & ColMap \\
%\midrule
%0.504 & 0.448 & 0.474 & ResNet50-RGB & G    & CDTB-G & G & 3$\times$G\\
%0.479 & 0.431 & 0.454 & ResNet50-RGB & G    & CDTB-G & G & ColMap\\
%0.383 & 0.354 & 0.368 & ResNet50-RGB & G    & CDTB-D & D & 3$\times$D\\
0.419 & 0.307 & 0.355             & ResNet50-RGB & D$^\dagger$  & CDTB-D & \fbox{D} & 3$\times$D\\
0.418 & 0.325 & \underline{0.366} & ResNet50-RGB & D$^\dagger$  & CDTB-D & \fbox{D} & ColMap\\
%0.438 & 0.401 & 0.419 & \bf{CDTB-LT} & ResNet50 & \checkmark &            &            &            & \checkmark \\
%0.432 & 0.314 & 0.363 & ResNet50-RGB & CDTB-D & D$^\dagger$  & D & 3$\times$D\\
%0.409 & 0.287 & 0.338 & ResNet50-RGB & CDTB-D & D$^\dagger$  & D & ColMap\\
         \bottomrule
    \end{tabular}}
    \caption{Cross-dataset results for the DepthTrack-ST test set.}
    \label{tab:exp_depth_2}
\end{table}

In the first experiment we investigated the contributions of depth-only training and depth-only fine-tuning of DiMP.
To further emphasize the limitation that ResNet-50 is pretrained with RGB from
ImageNet we denote it as "ResNet50-RGB".
The RGB DiMP is the original DiMP from its authors.
Depth variants of DiMP were trained with the generated depth data of LaSOT, GOT10k and COCO. 
% MOVED TO SETTINGS
%\song{For a fair comparison, we follow the same training settings as the DiMP.}
For sanity check we report results also for DiMP trained with different color channels of the same datasets.

\paragraph{CDTB-ST.}
The results are summarized in Table~\ref{tab:exp_depth} and provide the following findings:
{\bf 1)} off-the-shelf RGB DiMP is superior to single color channels (R/G/B) and depth as expected;
{\bf 2)} fine-tuning with the DepthTrack training set sequences systematically improves the results of all trackers and, in particular, depth tracking results are substantially improved (F-score jumps +20\% from 0.421 to 0.508);
{\bf 3)} training DiMP from the scratch using channel specific training data further improves the results of all trackers - G channel obtains better accuracy than the original RGB (F-score increases from 0.665 to 0.701) and the depth-only F-score further improves +9\% from 0.508 to 0.556.
% LETS COMPLETELY REMOVE IT AS IT IS CONFUSING
% {\bf 4)} \song{Colormap encoding (ColMap) is only harmful as compared to copying the single channel (3$\times$D).}
%{\bf 4)} \song{Colormap encoding (ColMap) performs on par with the single channel copying (3$\times$D).}

These results verify that by training and fine-tuning DiMP with depth data - even with synthetic - improves the results substantially.
The best color channel (Green) F-score improved from 0.638 to 0.701 (+10\%) and the depth (D)
from 0.421 to 0.556 (+32\%). Overall, after these optimizations the Depth-DiMP is only
16\% behind the original RGB DiMP. 

\paragraph{DepthTrack-ST.}
To validate the above findings the datasets were swapped and DepthTrack-ST test set was used in testing and CDTB in fine-tuning. 
The results are summarized in Table~\ref{tab:exp_depth_2}. 
For DepthTrack the sequences are selected to particularly challenge RGB tracking and this is evident from the results. 
Fine-tuning again improves the results 28\% (F-score from 0.288 to 0.370), but in this case generated data does not provide additional benefit (0.366) as the sequences are very different from test data. 
However, it should be noted that if there is no fine-tuning data available the generated depth still improves the results.

%----------FSCORE PER SEQ-----------------------------
\begin{figure}[t]
    \centering
    \includegraphics[width=0.85\textwidth]{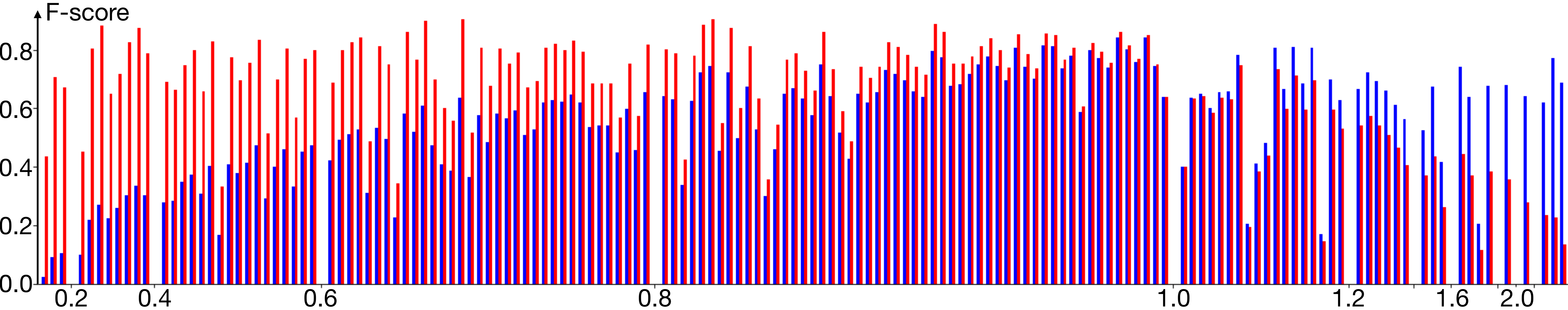}
    \caption{The plot of F-scores (Y-axis) and the ratios (X-axis) between the RGB DiMP tracker (red) and the Depth-DiMP tracker
    (D$^\dagger$-DepthTrack-3$\times$D) (blue)
    for each sequence in CDTB-ST.
    Ratios $<1.0$ represent
    RGB dominance (RGB performs better) and $>1.0$ D dominance.
    0.5 means that the RGB F-score is twice better than the corresponding D F-score and $2.0$ means that the depth is twice better.}
    \label{fig:fscore_cdtb_st}
\end{figure}
%-------------------------------------------------------
%--------------------ATTRIBUTES PLOTS------------------------
\begin{figure}[t]
    \centering
    \includegraphics[width=0.8\textwidth]{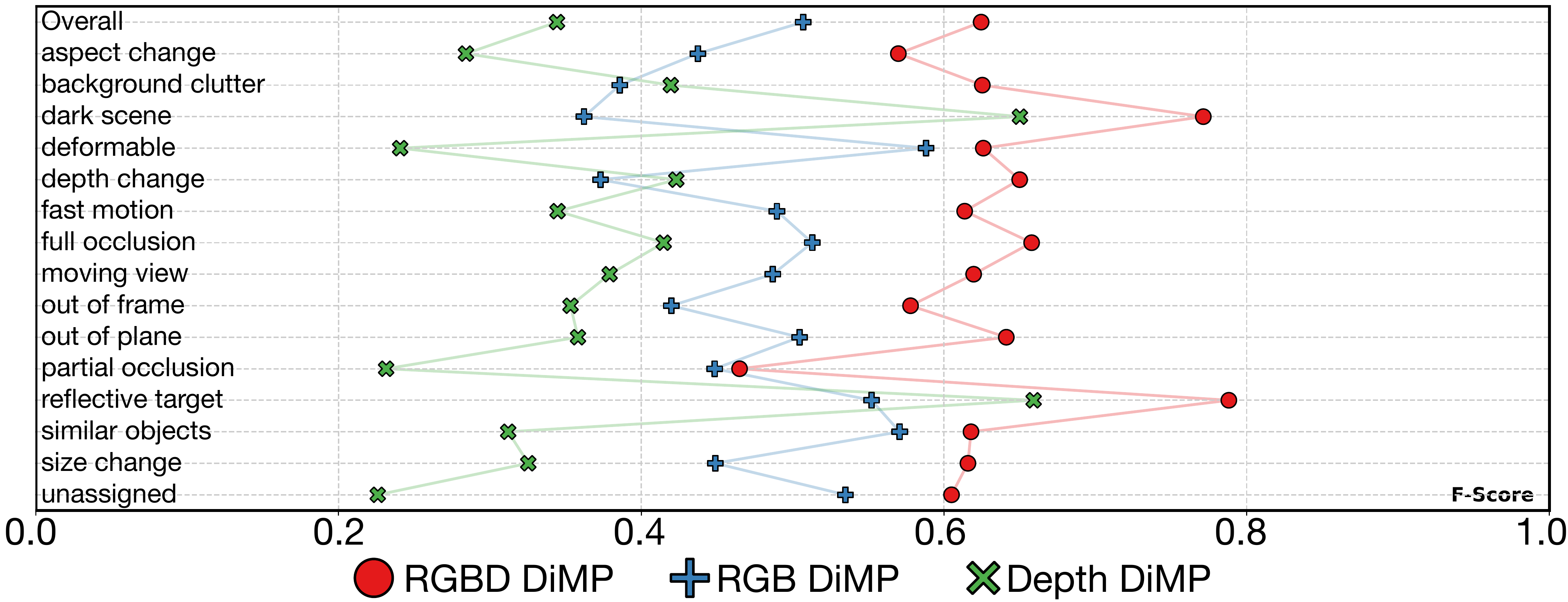}
    \caption{Visual attribute specific F-scores for RGB, D and RGBD (D$^\dagger$-CDTB-ColMap) DiMP variants on the DepthTrack-ST test set.}
    \label{fig:attributes_plot}
\end{figure}
%------------------------------------------------------------
%------------------------- Depth Superior ----------------------
\begin{figure}[t]
    \centering
    \includegraphics[width=0.75\textwidth]{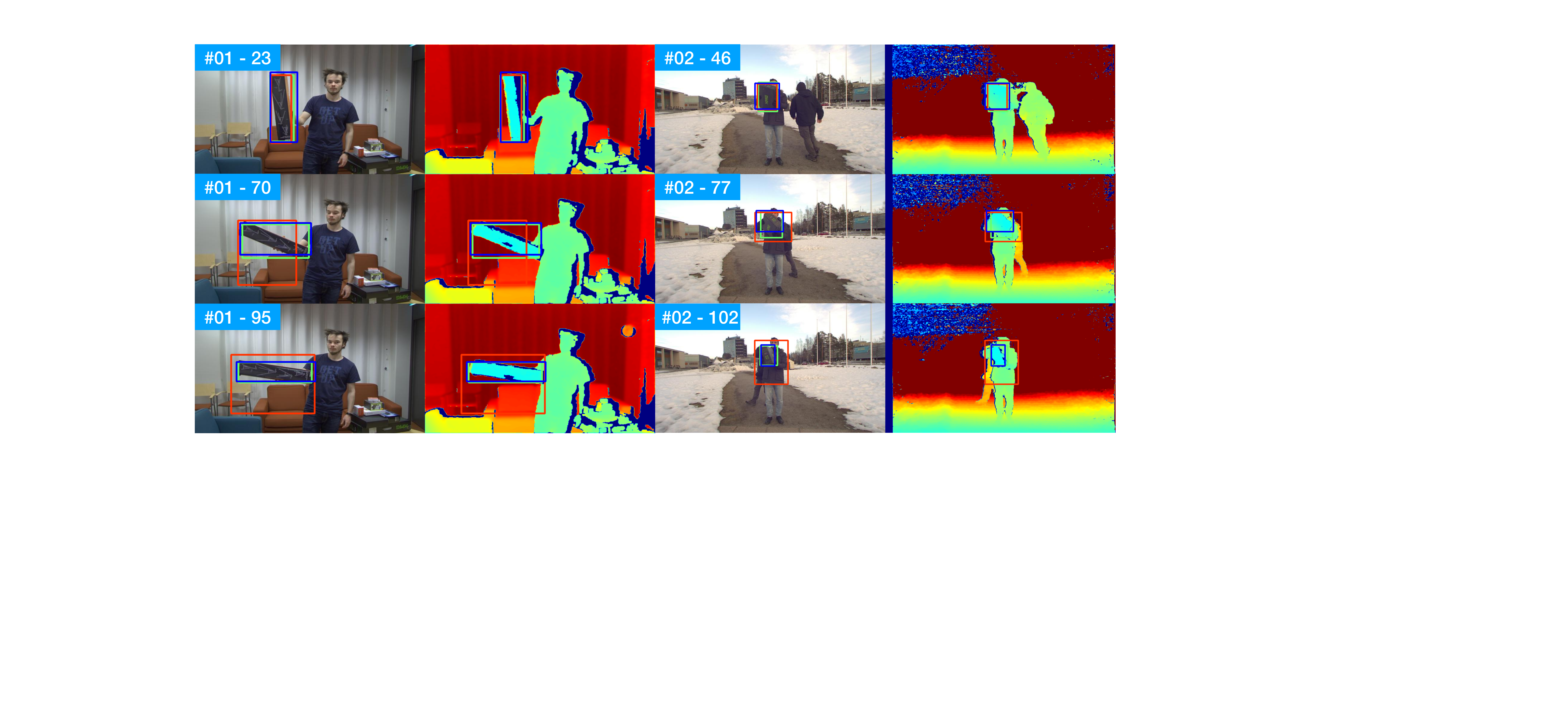}
    \caption{CDTB-ST example sequences
    where the Depth-DiMP tracker (D$^\dagger$-DepthTrack-3$\times$D) (blue boxes) is clearly superior to the RGB DiMP tracker (red boxes). Green boxes denote the groundtruth bounding boxes.
    {\em \#S-N} denotes the {\em N}-th frame of {\em S}-th Seq. in Table~\ref{tab:depth_superior_cdtb}. 
    }
    \label{fig:depth_superior}
\end{figure}
%------------------------------------------------------------
%--------------- RGB Superior -------------------------------
\begin{figure}[t]
    \centering
    \includegraphics[width=0.75\textwidth]{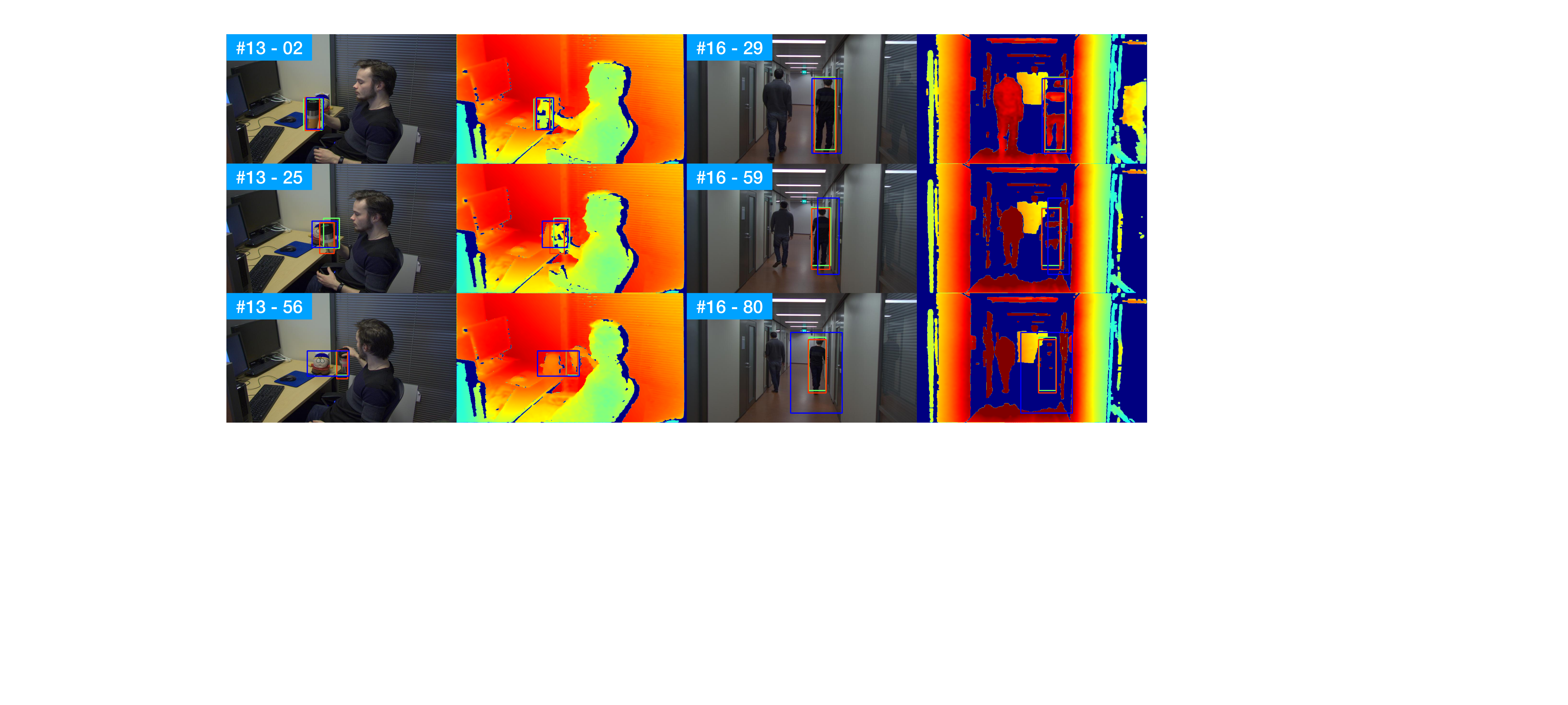}
    \caption{CDTB-ST example sequences
    where the RGB DiMP tracker (red boxes) is clearly superior to the Depth-DiMP tracker (D$^\dagger$-DepthTrack-3$\times$D) (blue boxes).}
    \label{fig:rgb_superior}
\end{figure}
%------------------------------------------------------------

%-----------------------------------------------------------
\begin{table}[t]
    \centering
    \resizebox{0.8\linewidth}{!}{
    \begin{tabular}{r l r r r c r r r}
    \toprule
        ~ & ~ & \multicolumn{3}{c}{D DiMP} & \multicolumn{3}{c}{RGB DiMP}\\
        {\em id} & {\em Seq. name} &  Pr & Re & F & Ratio & Pr & Re & F\\
        \hline
        01 & box\_room\_noocc\_3\_1 & 0.689 & 0.689 & 0.689 & $\leftarrow$ 5.1 & 0.135 & 0.135 & 0.135 \\
        02 & XMG\_outside\_2 & 0.771 & 0.771 & 0.771 & $\leftarrow$ 3.4 & 0.229 & 0.229 & 0.229 \\
        03 & boxes\_office\_occ\_1\_2 & 0.621 & 0.621 & 0.621 & $\leftarrow$ 2.6 & 0.235 & 0.235 & 0.235 \\
        04 & box\_darkroom\_noocc\_9\_1 & 0.643 & 0.643 & 0.643 & $\leftarrow$ 2.3 & 0.278 & 0.278 & 0.278 \\
        05 & boxes\_room\_occ\_1\_1 & 0.669 & 0.693 & 0.681 & $\leftarrow$ 1.9 & 0.354 & 0.360 & 0.357 \\
        06 & box\_darkroom\_noocc\_5\_1 & 0.677 & 0.679 & 0.678 & $\leftarrow$ 1.8 & 0.385 & 0.387 & 0.386 \\
        07 & box\_room\_occ\_1\_1 & 0.595 & 0.688 & 0.638 & $\leftarrow$ 1.7 & 0.346 & 0.403 & 0.372 \\
        08 & bag\_outside\_3 & 0.741 & 0.741 & 0.741 & $\leftarrow$ 1.7 & 0.444 & 0.444 & 0.444 \\
        09 & box\_room\_noocc\_2\_1 & 0.417 & 0.417 & 0.417 & $\leftarrow$ 1.6 & 0.262 & 0.262 & 0.262 \\
        10 & toy\_office\_occ\_1\_2 & 0.642 & 0.708 & 0.673 & $\leftarrow$ 1.5 & 0.541 & 0.367 & 0.437 \\
        
        \midrule
        
        11 & trophy\_room\_occ\_1\_2 & 0.083 & 0.103 & 0.092 & 7.7$\rightarrow$ & 0.653 & 0.770 & 0.706 \\
        12 & two\_mugs\_5 & 0.108 & 0.106 & 0.107 & 6.3$\rightarrow$ & 0.707 & 0.640 & 0.672 \\
        13 & thermos\_office\_occ\_1\_2 & 0.091 & 0.112 & 0.100 & 4.5$\rightarrow$ & 0.416 & 0.494 & 0.452 \\
        14 & box\_room\_noocc\_8\_1 & 0.220 & 0.220 & 0.220 & 3.7$\rightarrow$ & 0.804 & 0.804 & 0.804 \\
        15 & humans\_corridor\_occ\_2\_B\_4 & 0.246 & 0.298 & 0.270 & 3.3$\rightarrow$ & 0.882 & 0.882 & 0.882 \\
        16 & humans\_corridor\_occ\_2\_B\_1 & 0.160 & 0.372 & 0.224 & 2.9$\rightarrow$ & 0.538 & 0.825 & 0.651 \\
        17 & jug\_3 & 0.259 & 0.259 & 0.259 & 2.8$\rightarrow$ & 0.718 & 0.718 & 0.718 \\
        18 & bottle\_box\_2 & 0.232 & 0.434 & 0.303 & 2.7$\rightarrow$ & 0.846 & 0.808 & 0.827 \\
        19 & trashcan\_room\_occ\_1\_6 & 0.333 & 0.342 & 0.337 & 2.6$\rightarrow$ & 0.882 & 0.868 & 0.875 \\
        20 & bottle\_room\_occ\_1\_2 & 0.558 & 0.209 & 0.304 & 2.6$\rightarrow$ & 0.787 & 0.787 & 0.787 \\

        \bottomrule
    \end{tabular}}
    \caption{10 best CDTB-ST sequences for D and RGB DiMP trackers respectively. 
    Ratios are converted to indicate how many times better the other modality is as denoted by the arrow.
    }
    \label{tab:depth_superior_cdtb}
\end{table}
%---------------------------------------------------------------

%
\subsection{Sequence level analysis}
The results in Table~\ref{tab:exp_depth} show that the best depth-only
DiMP obtains F-score only 0.1 points behind the state-of-the-art
RGB DiMP (0.556 vs. 0.665). Since the results of RGB DiMP do not substantially improve
after fine-tuning (0.701 vs. 0.665) it is intriguing to
find whether
there are sequences particularly suitable or
unsuitable for depth-only tracking.

To investigate the complementary properties of RGB and D we computed the graph in Figure~\ref{fig:fscore_cdtb_st} where
all CDTB-ST sequences are sorted from left to right so that
for the leftmost images RGB is more dominant and for the rightmost
D is more dominant.
In the middle are sequences for which color and
depth perform comparably. In Figure~\ref{fig:fscore_cdtb_st}
for most of the
sequences RGB is better than D (ratio $<1.0$), but,
on the other hand, in 20\% of the sequences (36 out of 152)
the depth is better than RGB ($>1.0$) and in half of them
depth is clearly better. 

\paragraph{Example sequences.}
The 10 best and worst sequences for the depth-only tracking 
are listed in Table~\ref{tab:depth_superior_cdtb} and 
Figure~\ref{fig:depth_superior} shows two example sequences in which target shares similar texture and color with distractors (a person in these examples) and 3D rotates to change its 2D shape quickly. These changes finally lock the RGB tracker to wrong object. On the other hand,
the distinct depth of the targets make
their depth-only tracking easy.

Scenes for which depth-only fails
are illustrated in
Figure~\ref{fig:rgb_superior}. Two reasons can be observed: the target
and distractor share the same depth, but have different texture, or the target
is too far and the depth sensor fails to get measurements.

\paragraph{Visual attributes.}
To provide more insights about scenes in which depth is an important cue, 
we computed attribute specific F-scores for the DepthTrack-ST (Figure~\ref{fig:attributes_plot}).
The scores verify that there are scene types where depth outperforms RGB:
{\em background clutter},
{\em dark},
{\em depth change} and
{\em reflective} scenes.
RGB is better than D for the other scene types (attributes), but together RGB and D are
clearly superior in all of them.

\subsection{RGBD tracking}
We combined the optimized Depth-DiMP and RGB DiMP 
to RGBD-DiMP and compared its performance to the VOT2020-RGBD winners with the novel DepthTrack-ST test set.
The results in Table~\ref{tab:rgbd_comparision_VOT2021-RGBD_st} show that
the 2020 second, DDiMP, now wins the last year winner, ATCAIS,
but 
our end-to-end trained RGBD-DiMP obtains the best results with 16\% higher F-score than DDiMP.
%\song{Their performance on DepthTrack-ST test set are substantially lower than with the CDTB since more depth dominated scenes are introduced into DepthTrack
%but the tracking performances of SotA RGBD trackers heavily depend on RGB features,
%and depth is mainly used for occlusion detection.
%}
It should be noted that the short-term setting in these experiments is different from the long-term setting used in the VOT-RGBD
challenge.

%------------- RGBD Tracking ----------------------------------
\begin{table}[t]
    \centering
    \resizebox{0.6\linewidth}{!}{
    \begin{tabular}{l|c c c |c l}
    \toprule
    \multicolumn{1}{c|}{Tracker}   & Pr & Re & F & \multicolumn{1}{c}{Input}\\
    \midrule
    ATCAIS    & 0.566 & 0.444 & 0.498 & RGB+D\\
    DDiMP     & 0.607 & 0.478 & 0.535 & RGB+D\\
    CLGS\_D   & {\bf 0.710} & 0.367 & 0.484 & RGB+D\\
    Siam\_LTD & 0.433 & 0.389 & 0.410 & RGB+D \\
    \midrule
    DiMP-RGBD & \underline{0.639} & {\bf 0.611} & {\bf 0.625} & RGB+ColMap\\
    DiMP-RGBD & 0.610 & \underline{0.567} & \underline{0.588}  & RGB+3$\times$D\\
    \bottomrule
    \end{tabular}}
    \caption{Comparison of the four best VOT2020-RGBD trackers~\cite{Vot2020} and the RGBD data trained RGBD-DiMP with the DepthTrack-ST test set.}
    \label{tab:rgbd_comparision_VOT2021-RGBD_st}
\end{table}
%---------------------------------------------

%---------------------------------------------
\section{Conclusion}
\label{sec:conclusion}
We experimentally verified that there are tracking scenes for which depth provides dominant cue and where depth-only trackers are superior to state-of-the-art RGB trackers. Moreover, we showed that the end-to-end trained RGBD-DiMP using generated and real depth and RGB data is clearly superior to state-of-the-art RGBD trackers and does not need additional heuristics. Our data and code will be published to facilitate fair comparisons and more work on depth-only object tracking.

%---------------------------------------------
\section*{Acknowledgements}
We thank VOT guru Matej Kristan for the original idea of using generated depth in training.

\bibliography{egbib}
\end{document}